\documentclass[10pt,twocolumn,letterpaper]{article}
\usepackage{url}

\usepackage[pagenumbers]{cvpr} 

\usepackage[dvipsnames]{xcolor}

\definecolor{cvprblue}{rgb}{0.21,0.49,0.74}
\usepackage[pagebackref,breaklinks,colorlinks,citecolor=cvprblue]{hyperref}

\def\paperID{1754} 
\def\confName{CVPR}
\def\confYear{2024}

\title{Triamese-ViT: A 3D-Aware Method for Robust Brain Age Estimation from MRIs}

\author{Zhaonian Zhang \\
Lancaster University\\
Lancaster LA1 4YZ, England\\
{\tt\small z.zhang47@lancaster.ac.uk}
\and
Richard Jiang \\
Lancaster University\\
Lancaster LA1 4YZ, England\\
{\tt\small r.jiang2@lancaster.ac.uk}
}

\begin{document}
\maketitle
\begin{abstract}
The integration of machine learning in medicine has significantly improved diagnostic precision, particularly in the interpretation of complex structures like the human brain. Diagnosing challenging conditions such as Alzheimer's disease has prompted the development of brain age estimation techniques. These methods often leverage three-dimensional Magnetic Resonance Imaging (MRI) scans, with recent studies emphasizing the efficacy of 3D convolutional neural networks (CNNs) like 3D ResNet. However, the untapped potential of Vision Transformers (ViTs), known for their accuracy and interpretability, persists in this domain due to limitations in their 3D versions. This paper introduces Triamese-ViT, an innovative adaptation of the ViT model for brain age estimation. Our model uniquely combines ViTs from three different orientations to capture 3D information, significantly enhancing accuracy and interpretability. Tested on a dataset of 1351 MRI scans, Triamese-ViT achieves a Mean Absolute Error (MAE) of 3.84, a 0.9 Spearman correlation coefficient with chronological age, and a -0.29 Spearman correlation coefficient between the brain age gap (BAG) and chronological age, significantly better than previous methods for brian age estimation. A key innovation of Triamese-ViT is its capacity to generate a comprehensive 3D-like attention map, synthesized from 2D attention maps of each orientation-specific ViT. This feature is particularly beneficial for in-depth brain age analysis and disease diagnosis, offering deeper insights into brain health and the mechanisms of age-related neural changes.
\end{abstract}    
\section{Introduction}
\label{sec:intro}

\begin{figure*}[htbp]
\centering
\includegraphics[scale=0.7]{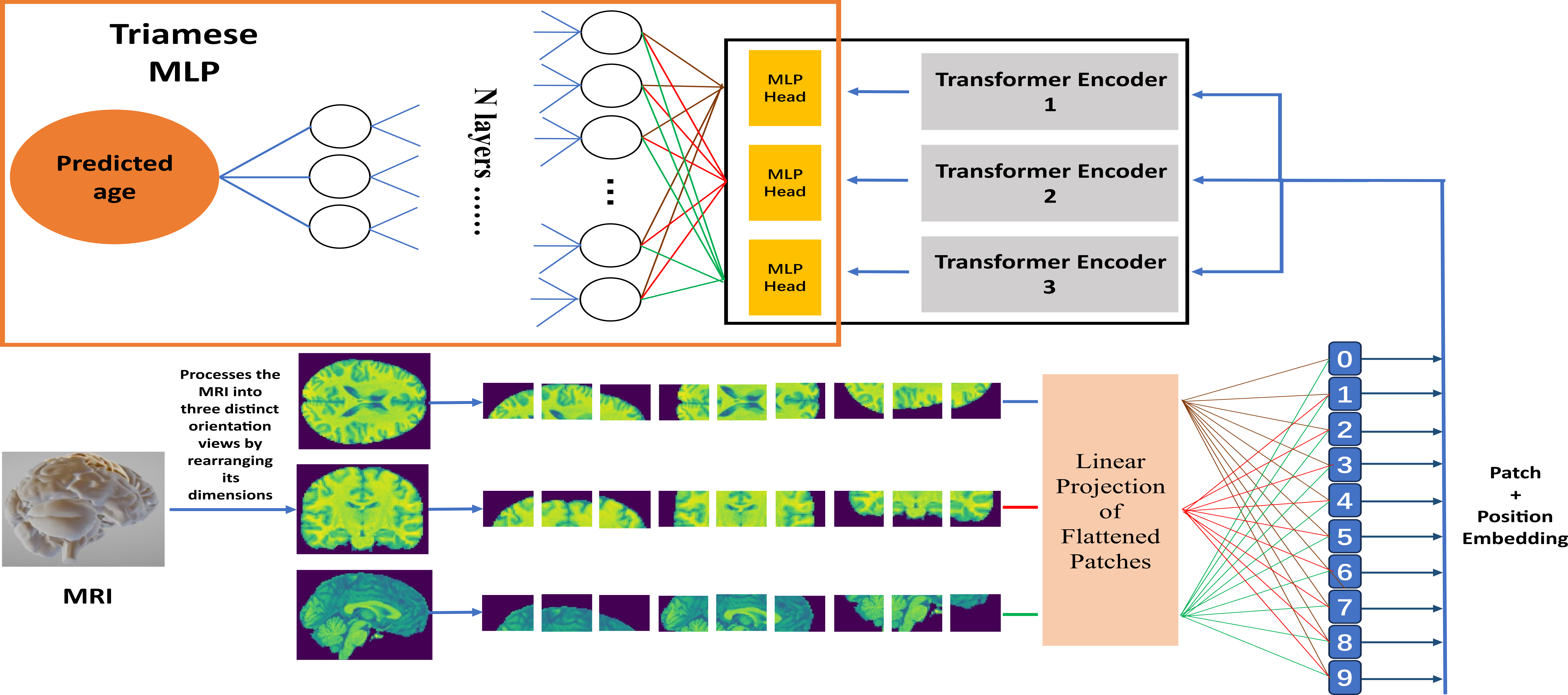}
\caption{\textbf{The structure of Triamese-ViT.} We reshape MRI scans into three distinct viewpoints, dividing each into fixed-size patches. These patches are then linearly embedded, enhanced with position embeddings, and subsequently inputted into a standard Transformer encoder. The encoder's output is directed through MLP Heads to generate three separate predictions. These predictions are then integrated using the Triamese MLP, culminating in the final result.}
\label{figure 1}
\end{figure*}

Aging naturally impacts all body parts, including the brain, which is particularly sensitive to age-related changes, heightening the risk of diseases like Alzheimer's. Traditional diagnostic methods for brain diseases are often slow and heavily reliant on subjective clinical judgment~\cite{avants2008symmetric,liu2015low}, leading to potential delays or inaccuracies in diagnosis. Such shortcomings can be critical, risking the loss of crucial treatment time and possibly exacerbating the patient's condition.

Recent advances in machine learning, particularly deep learning, have revolutionized brain diagnostics~\cite{cole2017predicting,feng2020estimating,armanious2021age,zhang2022robust}. Deep learning's ability to estimate brain age from images is pivotal in detecting age-related diseases~\cite{cole2017predicting,beheshti2018association,gaser2013brainage}. The brain age gap (BAG), the disparity between estimated and actual brain age, is a crucial indicator~\cite{cole2017predicting}. A younger-appearing brain indicates health, while an older-looking brain may signal conditions like Alzheimer's~\cite{beheshti2018association}, psychosis~\cite{chung2018use}, mild cognitive impairment~\cite{gaser2013brainage}, or depression~\cite{han2021brain}. Enhancing brain age estimation algorithms is essential, facilitating early disease detection and offering patients hope through improved treatment prospects.

Current brain age estimation largely depends on convolutional neural networks (CNNs) trained on either 3D MRI scans~\cite{cole2017predicting} or 2D slices from these scans~\cite{hong2020brain,bellantuono2021predicting}. While CNNs excel in image processing and can utilize full 3D MRI data for comprehensive predictions, they struggle with global feature representation due to their focus on small, local pixel groups~\cite{hu2022sqet}. This results in a loss of critical details, particularly in the analysis of the brain's complex structures. Moreover, CNNs' lack of transparency poses challenges in Explainable AI, making their predictions difficult to interpret in medical diagnostics, especially when pinpointing specific brain abnormalities~\cite{alzubaidi2021review}.

The Vision Transformer (ViT) presents a notable advantage over traditional CNNs in image detail analysis~\cite{dosovitskiy2020image}. By segmenting images into patches and transforming each via a convolutional layer into a high-dimensional space, ViT effectively extracts intricate details and understands inter-part interactions~\cite{khan2022transformers}. Its attention map feature provides insight into its focus and prediction rationale. However, ViT has limitations: it may overlook broader, global contexts~\cite{tanveer2023deep} and is primarily suited for 2D images~\cite{al2023vision}. For brain age prediction, since researchers often use flat 2D slices of MRI scans as input for ViTs~\cite{he2021global}, although this approach can greatly reduce the time of training and prediction, it doesn't take into account the full 3D structure of the brain, leading to a loss of important depth information. 

To address the need for both detailed analysis and understanding of interactions across different areas within an image, we introduce a new model, Triamese-ViT, whose design is a first in the realm of brain age estimation. This model is inspired by the Siamese Networks, which has been successfully applied to image classification, detection and comparison \cite{cui2022joint,zhang2023similarity,zeng2022siampcf}. The novelty of our model is that it adeptly transforms 3D imaging challenges into a 2D analytical framework, effectively preserving the rich information contained in three-dimensional images, capturing the relationship between different areas in the brain very well, and has the same fast process speed as 2D images at the same time. 

A standout feature of our model is its capability to generate 3D-like attention maps, offering profound interpretability. We meticulously compared these maps with conventional Explainable AI (XAI) methods and corroborated with findings from relevant medical research~\cite{shine2023impact,smajic2022single,russo2022age,xiong2022altered,wang2022roles}, affirming their scientific validity, logical coherence, and precision. This synergy of advanced imaging analysis and interpretability positions our model as a significant tool in medical diagnostics and research, especially in areas demanding high fidelity in three-dimensional medical data understanding. 

We trained and tested our model on MRI scans from 1351 healthy individuals. When compared with the state-of-the-art algorithms, Triamese-ViT not only showed superior predictive accuracy but also demonstrated improvements in fairness and interoperability. Our model achieves impressive results that mean absolute error (MAE) of 3.87, a 0.93 Spearman correlation coefficient when comparing predicted brain age to chronological age, and a -0.29 Spearman correlation coefficient between the chronological age and the brain age gap (the absolute value of the gap between the predicted and actual age), significantly better than previous methods for brain age estimation.

In addition, we subjected our model to Explainable AI (XAI) techniques, such as 3D occlusion analysis, to validate the high-correlation areas identified by the model. These areas matched those highlighted by the attention maps produced by Triamese-ViT, reinforcing the validity of our model’s interpretive output.

\section{Related work}
This section will introduce the background which is related to our project. While rapid advancements in brain age estimation have been made possible through deep learning, the field still faces significant challenges. One key issue is the lack of comprehensive research into which specific brain regions most significantly impact age estimation~\cite{tanveer2023deep}. Limited studies in this area have indicated the presence of ageism, as accuracy varies across different age groups~\cite{tanveer2023deep,zhanguser}. Additionally, deep learning models, often described as 'black boxes,' lack transparent processes for mapping input components to output values~\cite{BARREDOARRIETA202082}. This limitation is particularly evident when compared to the more interpretable, yet less effective, algorithmic models like decision trees. The emergence of post-hoc explainability methods, such as SHAP and LIME\cite{lombardi2021explainable}, marks a crucial development in addressing this issue, though their application in brain age estimation remains limited. Future research should prioritize the inclusion of these explainability methods, focusing not just on performance, but also on reliability and trustworthiness for clinical application, to ensure that models are both effective and comprehensible in a medical context~\cite{holzinger2022information}.

Convolutional neural networks (CNNs) are very popular to be applied for the field of brain age estimation. Since 2017, CNNs have been at the forefront, favored for their ability to autonomously extract features and deliver high-accuracy results~\cite{bellantuono2021predicting,pardakhti2020brain,jiang2020predicting,hong2020brain}. Cole et al.~\cite{cole2017predicting} demonstrated this by training a 3D CNN on MRI samples, finding that models trained on gray matter outperformed those trained on white matter. To enhance efficiency, researchers have turned to 2D MRI slices, with ~\cite{lam2020accurate} using recurrent neural networks (RNN) to understand the connections between slices, offering a balance between performance and computational demand. Then, the advanced and special CNN structure such as SFCN, ResNet, and DenseNet also highlighted their strength in this area~\cite{ning2021improving,mouches2022multimodal,wood2022accurate}. When the transformer generally become a hot topic, since it can help pay more attention to details of the images, it is very suitable for use in the medical field. For instance, the Global–Local Transformer~\cite{he2021global} melds the strengths of CNNs with transformers, utilizing 2D MRI slices to capture both local and global information in the images. Cai et al.~\cite{cai2022graph} furthered this approach by leveraging a graph transformer network in a multimodal method, efficiently harnessing both global and local features for brain age estimation.

Since we will use XAI techniques to prove our model's ability to interpretability, we will introduce the background of XAI. Despite its wide-ranging utility, AI often operates as a "black box," with complex models and a multitude of parameters obscuring the decision-making process~\cite{langer2021we,langer2021explainability,doshi2017towards,ali2023explainable}. This lack of transparency is particularly critical in sensitive fields such as finance~\cite{cao2022ai} and healthcare~\cite{caruana2015intelligible}, where understanding AI's reasoning is crucial, so this need boosts the development of XAI. 

XAI methods generally fall into two categories~\cite{longo2023explainable,das2020opportunities}: backpropagation-based and perturbation-based. Backpropagation-based XAI involves algorithms that provide insight during the backpropagation stage of neural network processing, typically by using derivatives to produce attribution maps like class activation or saliency maps. On the other hand, perturbation-based approaches modify the input features in various ways—through occlusion, substitution, or generative techniques—to observe changes in the output. Our project employs occlusion analysis, a perturbation-based technique, to validate our model's interpretability.

Our Triamese-ViT model is an adaptation of Siamese Networks, a class of neural networks comprising twin subnetworks that share weights while processing distinct inputs. Renowned for their performance, Siamese Networks have been notably utilized in various fields. Zeng et al.\cite{zeng2022siampcf} employed this network structure to develop an anchor-free tracking method, leading in scale variance for video attribute analysis. Zhang et al.\cite{zhang2023similarity} applied these networks for regression tasks on physicochemical datasets, achieving high accuracy. Additionally, SiamCAR, created by Cui et al.~\cite{cui2022joint} using the Siamese architecture, excelled in real-time visual tracking of generic objects.

\section{Method}
\subsection{Proposed Triamese-ViT}
In this section, we will introduce the structure of Triamese-ViT. As we show in Fig.\ref{figure 1}, the idea of Triamese-ViT is inspired by ViT ~\cite{dosovitskiy2020image} and Siamese Networks. The input of Triamese-ViT are 3D MRIs, we called the images $I$$\in$$R^{H\times W\times C}$, here H, W, C are image height, width and number of channels. Then we reshape the image $I$ into 3 different viewpoints, $I$$\to$($I_x$,$I_y$,$I_z$), $I_x$$\in$$R^{H\times W\times C}$,$I_y$$\in$$R^{H\times C\times W}$,$I_z$$\in$$R^{W\times H\times C}$.
Firstly, we focus on $I_x$, we change it to a sequence of flattened 2D patches called $I_{x,p}$$\in$$R^{N\times(p^{2}\cdot C)}$, the patches are squares of length $P$. So the number of patches is $N=H\times W/P^{2}$.
In all of the layers in the transformer encoder, vectors with dimension D will be processed as the object, so we need to map $I_x$ to D dimensions with a trainable linear projection. The formulate is shown below:
\begin{equation}
z_{x,0}=Concat(I_{x,class};I_{x,p}^{1}E;I_{x,p}^{2}E;...;I_{x,p}^{N}E)+E_{pos}
\label{1}
\end{equation}
In Eq. \ref{1}, $I_{x,class}$ means a learnable token, in other words, the class token, which is added into ViT, this method is similar to ~\cite{devlin2018bert}. Note that $I_{x,class}$ will finally output from the Transformer Encoder as $z_{x,L}^{0}$, which represents the image representation $P$ (Eq. \ref{7}). $E \in R^{(p^{2}\cdot C)\times D}$ means Linear Projection, $Concat$ means token concatenation and $E_{pos} \in R^{(N+1)\times D}$ represents positional information which is added to each token embedding. We use the same pre-processing methods on $I_y$ and $I_z$, and we can get $z_{y,0}$ and $z_{z,0}$.

Now, the original data is processed into three suitable matrices $z_{x,0},z_{y,0},z_{z,0} \in R^{(N+1)\times D}$, then we feed them to the transformer encoder. Each Transformer encoder is a multi-layered system where the input sequentially passes through several key components in each layer: it first encounters a Layer Normalization (LN), followed by a Multi-Head Attention mechanism, another Layer Normalization, and then a Multi-Layer Perceptron (MLP). The Multi-Head Attention (MSA) operates by conducting parallel attention calculations across multiple 'heads', diversifying the focus and allowing for a richer understanding of the input data.
\begin{equation}
[Q,K,V]=FC(z_{x,0})
\label{2}
\end{equation}
Here, $Q\in R^{(N+1)\times d}$,$K\in R^{(N+1)\times d}$,$V\in R^{(N+1)\times d}$ represent Query, Keyword and Value. We assume the number of heads in MSA is n, and $D=n\times d$. Each 'head' independently processes the input, allowing the model to simultaneously attend to information from different representation subspaces at different positions.
\begin{equation}
head_{i}=softmax(\frac{Q_{i}K_{i}^{\mathsf{T}}}{\sqrt{d}}V_{i})
\label{3}
\end{equation}
\begin{equation}
MSA(z_{x,0})=Concat(head_{1},head_{2},...,head_{n})
\label{4}
\end{equation}

\noindent Let $z_{x,0}$ be the input of the first layer Transformer Encoder, so the feedforward calculation in the Encoder is written as:
\begin{equation}
z_{x,l}^{'}=MSA(LN(z_{x,l-1}))+z_{x,l-1}
\label{5}
\end{equation}
\begin{equation}
z_{x,l}=MLP(LN(z_{x,l}^{'}))+z_{x,l}^{'}
\label{6}
\end{equation}
The $l\in [1,2,...L]$. The outputs from each Transformer Encoder are channeled into an MLP (Multi-Layer Perceptron) head, comprising a hidden layer followed by an output layer, to generate the final prediction for each view. We denote the prediction from the $I_{x}$ (the first view) as $P_{x}$. Applying the same procedure to the $I_{y}$ and $I_{z}$ views, we obtain two additional predictions, $P_{y}$ and $P_{z}$, corresponding to these orientations.

In the final stage, these three view-based predictions ($P_{x}$, $P_{y}$, and $P_{z}$) are input into the Triamese MLP. This step integrates the insights from all three views to produce the model's comprehensive final prediction.
\begin{equation}
P= MLP(P_{x},P_{y},P_{z})
\label{7}
\end{equation}

\noindent We experimented with various strategies to integrate the outputs from the three viewpoint-specific ViTs. These strategies included averaging the results, selecting the best-performing output, and combining the outputs through an MLP. Our experiments demonstrated that processing the results through an MLP yielded the most effective performance. The loss function used in the training is the Mean Squared Error (MSE) between the predicted age $P_{i}$ and chronological age $C_{i}$.
\begin{equation}
Loss=\frac{\sum_{i=1}^{n}(P_{i}-C_{i})^{2}}{n}
\label{8}
\end{equation}

\subsection{Occlusion Sensitivity Analysis of Triamese-ViT}
To demonstrate the interpretability of our Triamese-ViT model, we employ Occlusion Sensitivity Analysis to generate saliency maps, which we then compare to the model's inherent attention maps. In this section, we will delve into the specifics of how Occlusion Sensitivity Analysis is conducted.

This analysis method systematically obscures different parts of the input data — in this case, regions of the brain within MRI images — to assess their impact on the model's output. By applying a cube-shaped mask (7x7x7 in our case) that sets the covered voxels to zero, and moving this mask throughout the entire volume of the brain without overlap, we can monitor changes in the model's predictions. The difference in prediction accuracy, quantified by the Mean Absolute Error (MAE) with and without occlusion, indicates the relative importance of each region.

\begin{figure*}[htbp]
\centering
\includegraphics[scale=0.8]{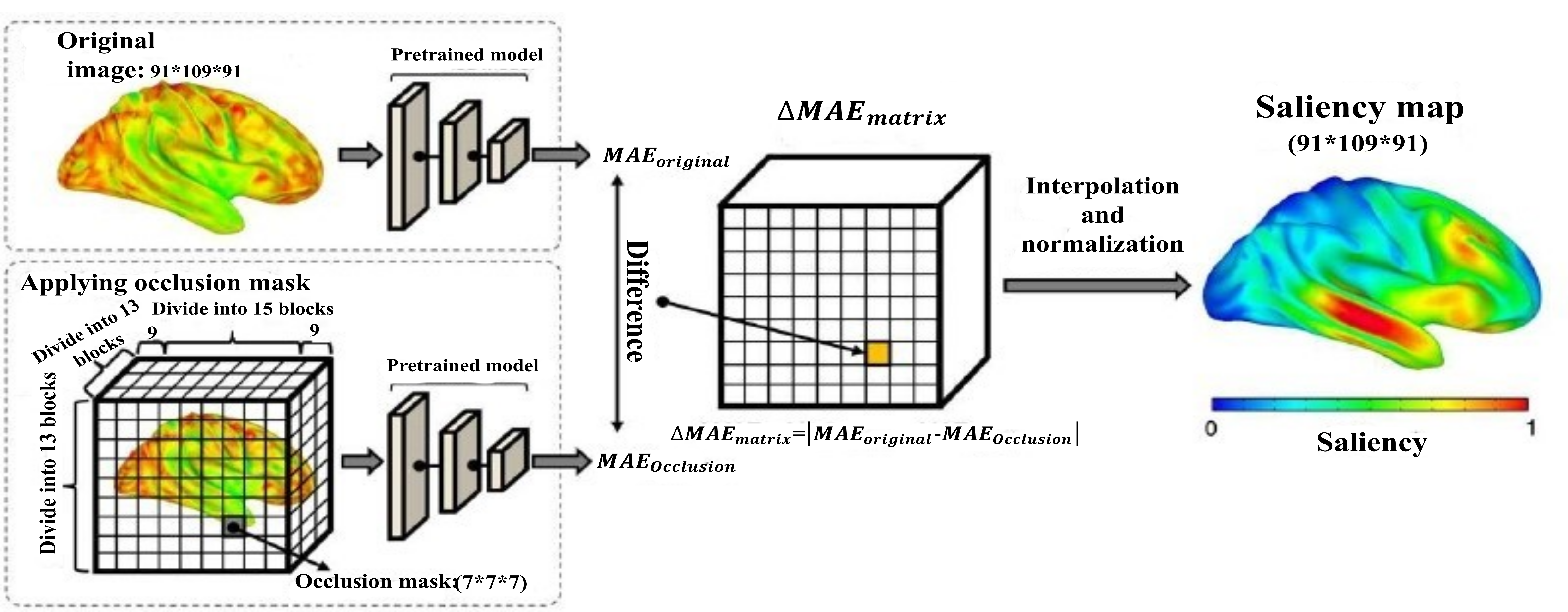}
\caption{Illustration of the framework for occlusion analysis.}
\label{figure 3}
\end{figure*}

By mapping these changes, we create a saliency map that highlights critical areas the model focuses on for its predictions. Comparing these saliency maps with the attention maps produced by Triamese-ViT provides a dual perspective on the model's decision-making process, offering a clear view of how it interprets the brain images to estimate age, and proving the ability of interpretability of the Triamese-ViT. The illustration is shown in Fig.\ref{figure 3}

\section{Experiments}
\subsection{Dataset}
Our research utilized the IXI \url{https://brain-development.org/ixi-dataset/} and ABIDE datasets \url{https://fcon_1000.projects.nitrc.org/indi/abide/}, encompassing 1351 brain MRI scans from healthy individuals aged 6 to 90 years. The data allocation was as follows: 70$\%$ for training the Triamese-ViT model, 15$\%$ for validation, and the remaining 15$\%$ for testing to evaluate model performance.

The MRI scans in our datasets underwent a standardized preprocessing routine using FSL 5.10~\cite{jenkinson2012fsl}. This process involved several key steps: nonlinear registration to the standard MNI space, brain extraction~\cite{smith2002fast}, and normalization of voxel values within the brain area (achieved by subtracting the mean and dividing by the standard deviation of these values). Post-preprocessing, all MRI scans were unified to a voxel size of $91\times 109\times 91$ and an isotropic spatial resolution of 2mm.

\subsection{Experimental Settings}
In our experiments, all models used the ADAM optimizer on PyTorch, with a 0.001 initial learning rate, $10^{-6}$ decay rate, and $\beta_{1} = 0.9$, $\beta_{2} = 0.999$. He initialization and L2 norm weight regularization (weight $5\times 10^{-4}$) were employed. Batch size was 100, with regularization weights $\lambda_{1}$ and $\lambda_{2}$ both at 10, a decision informed by the models’ performance on the validation set, and a manual seed of 3407 due to research~\cite{picard2021torch}.

Data augmentation on MRIs, in order to further mitigate the risk of overfitting, involved 50$\%$ probability of spatial transformations like 3D translation (up to 10 voxels), rotation (-20° to 20°), and random flips. For Triamese-ViT, each ViT had a $7\times7$ patch size, 768 embedding dimension, 12 attention heads, 10 layers, and 0.1 dropout rate. The MLP inside each ViT had one hidden layer with 3072 dimensionality. The Triamese MLP's 9 layers formed a pyramid sequence (3, 128, 256, 512, 1024, downscaling back to 3), synthesizing data from the ViTs into a single output.

\subsection{Performance Evaluation of Age Estimation}
In evaluating the performance of our age estimation model, we employ three key metrics. First is the mean absolute error (MAE) between the predicted age and chronological age, it is a direct measure of the model's accuracy; a lower MAE indicates higher accuracy. The MAE is calculated as follows:
\begin{equation}
MAE=\frac{\sum_{i=1}^{n}\left|(P_{i}-C_{i})\right|}{n}
\label{9}
\end{equation}

The second metric is the correlation coefficient (r), which is computed as the Spearman correlation between the predicted ages and the chronological ages. A higher value of 'rp signifies better model performance. The Spearman correlation is computed using the following formula:
\begin{equation}
r=\frac{\sum_{i=1}^{n}(P_{i}-\Bar{P})(C_{i}-\Bar{C})}{\sqrt{\sum_{i}^{n}(P_{i}-\Bar{P})^{2}\sum_{i}^{n}(C_{i}-\Bar{C})^{2}}}
\label{10}
\end{equation}

The last metric is the Spearman correlation between the chronological age and BAG (the absolute value of the gap between the predicted and actual age) (rp). This metric evaluates the fairness of the model, particularly checking for age bias in predictions. A higher correlation in this context suggests more pronounced ageism. The formula for this metric is: 
\begin{equation}
G_{i}=\left|P_{i}-C_{i}\right|
\label{11}
\end{equation}
\begin{equation}
rp=\frac{\sum_{i=1}^{n}(G_{i}-\Bar{G})(C_{i}-\Bar{C})}{\sqrt{\sum_{i}^{n}(G_{i}-\Bar{G})^{2}\sum_{i}^{n}(C_{i}-\Bar{C})^{2}}}
\label{12}
\end{equation}

\subsection{Comparison With State-of-the-Art Algorithms for Brain Age Estimation}
In this section, we'll present a comparative analysis of our Triamese-ViT model against various other state-of-the-art brain age estimation algorithms. This comparison, based on their performance on our dataset, aims to highlight the advancements and superior capabilities of our proposed model.
\begin{table}[b]
    \centering
    \begin{tabular}{p{4cm}lll}
        \hline
        Algorithm & MAE & r & rp \\
        \hline
        ScaleDense~\cite{cheng2021brain} & 3.92 & 0.92 & 0.39 \\
        5-layer CNN~\cite{couvy2020ensemble}   & 4.51& 0.78 & 0.45 \\
        ResNet~\cite{cole2017predicting}  & 4.01     & 0.83 & -0.31\\
        VGG19~\cite{huang2017age} & 4.07     & 0.7 & 0.5\\
        VGG16~\cite{huang2017age} & 5.24     & 0.6 &0.42\\
        Global-Local Transformer~\cite{he2021global}&4.66     & 0.78 & -0.32\\
        Efficient Net~\cite{poloni2022deep}& 4.56    & 0.89 &-0.4\\
        Our Triamese-ViT & $\mathbf{3.87}$ & $\mathbf{0.93}$ & $\mathbf{-0.29}$\\
        \hline
    \end{tabular}
    \caption{ The details of tested algorithms’ performance. Since the input of Global-Local Transformer should be 2D image, we extract 2D slices around the center of the 3D brain volumes in the axial as input, which is the same process method as ~\cite{he2021global}. Other algorithms' input are 3D MRIs with dimensions (91,109,91). Our u-DemAI has consistently achieved the best among all measures.}
    \label{Table 1}
\end{table}

Table \ref{Table 1} in our study provides a detailed comparison of our Triamese-ViT model with seven other models, encompassing both classic and contemporary approaches to brain age estimation. The comparison includes five well-known 3D CNN-based models: ScaleDense, a 5-layer CNN, ResNet, VGG16, and VGG19. Additionally, we evaluated against two other high-performing methods: the Global-Local Transformer, which is trained on 2D slices of the brain, and Effificent Net, known for its ensemble structure.

According to Table \ref{Table 1}, our Triamese-ViT model leads in Mean Absolute Error (MAE) performance with a score of 3.87. ScaleDense also performs impressively, achieving an MAE of 3.92, followed by ResNet with 4.01. The highest MAE, indicating the least accuracy, was recorded for VGG16 at 5.24.

In terms of the Spearman Correlation between predicted and chronological ages, our Triamese-ViT tops the list with a correlation of 0.93. ScaleDense is close behind with a correlation of 0.92, and ResNet follows with 0.83. VGG16 trails in this metric as well, showing a correlation of only 0.6.

Regarding the Spearman correlation between the Brain Age Gap (BAG) and chronological age, which reflects the fairness of the model, Triamese-ViT demonstrates the best outcome with a -0.29 correlation, indicating lower age bias. ResNet is next with a -0.31 correlation. On the other hand, VGG19 shows the most pronounced age bias with a 0.5 positive correlation.

Overall, this comparison underscores the effectiveness of Triamese-ViT in brain age estimation, both in terms of accuracy and fairness, when benchmarked against other leading methods in the field.

\subsection{Ablation Experiments}
In this part of our study, we conduct ablation experiments to explore and justify the design choices in the structure of Triamese-ViT. First of all, we specifically focus on the number of layers in the Triamese MLP. While keeping all other variables constant, we varied the number of MLP layers and observed their impact on the model's performance.

\begin{figure}[htbp]
\centering
\includegraphics[scale=0.32]{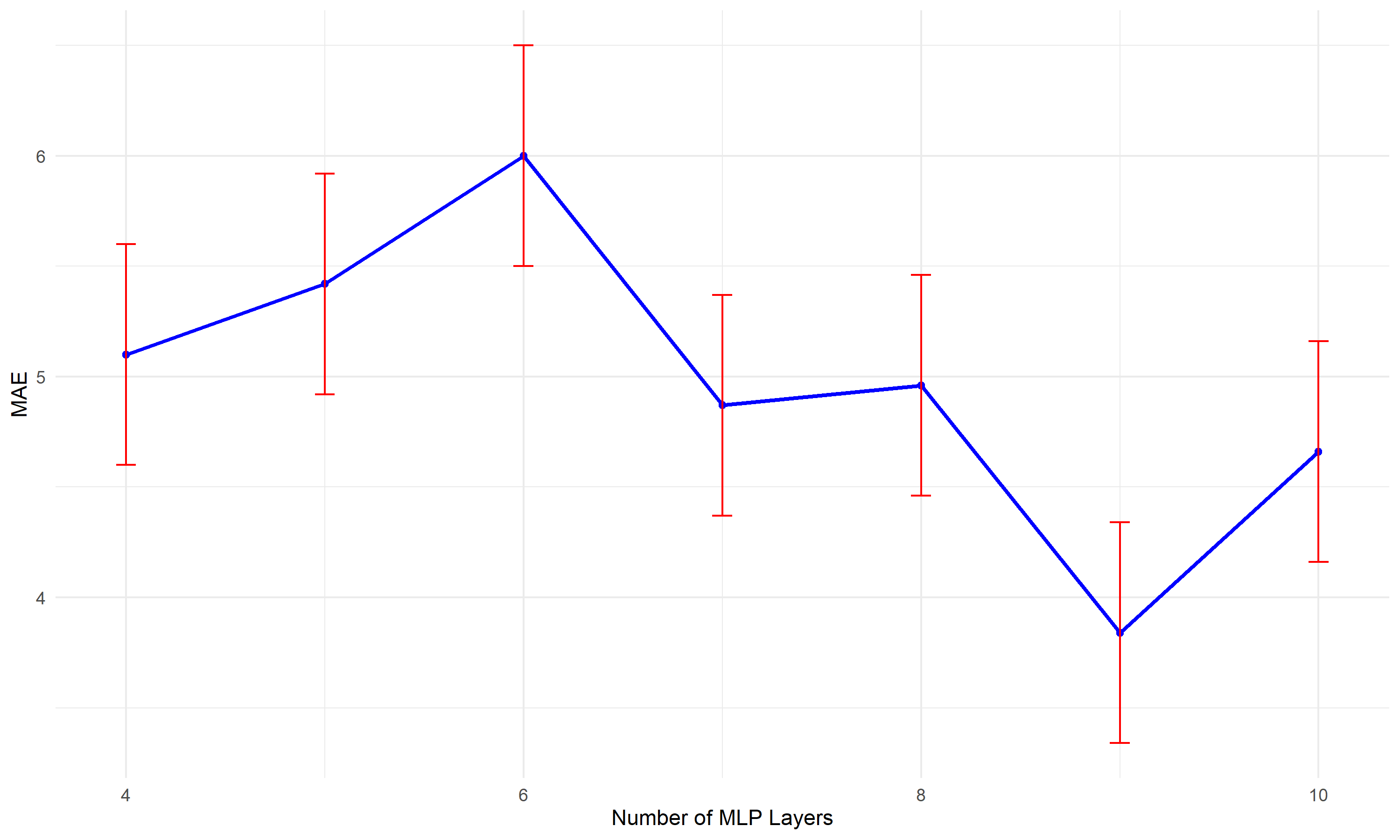}
\caption{The impact of the number of MLP layers in Triamese-Encoder}
\label{figure 4}
\end{figure}

The findings depicted in Fig.~\ref{figure 4} show a distinctive trend in the Mean Absolute Error (MAE) relative to the MLP layers in Triamese MLP. The MAE initially rises when increasing layers from 4 to 6, then decreases after 6 layers, reaching a minimum at 9 layers before rising again at 10 layers. This indicates an optimal layer count for balancing model complexity and accuracy. The observed MAE variation with different layer counts underscores the intricate relationship between model depth and performance, emphasizing the need for precise architectural tuning in the model.

\begin{figure}[!t]
\centering
\includegraphics[scale=0.32]{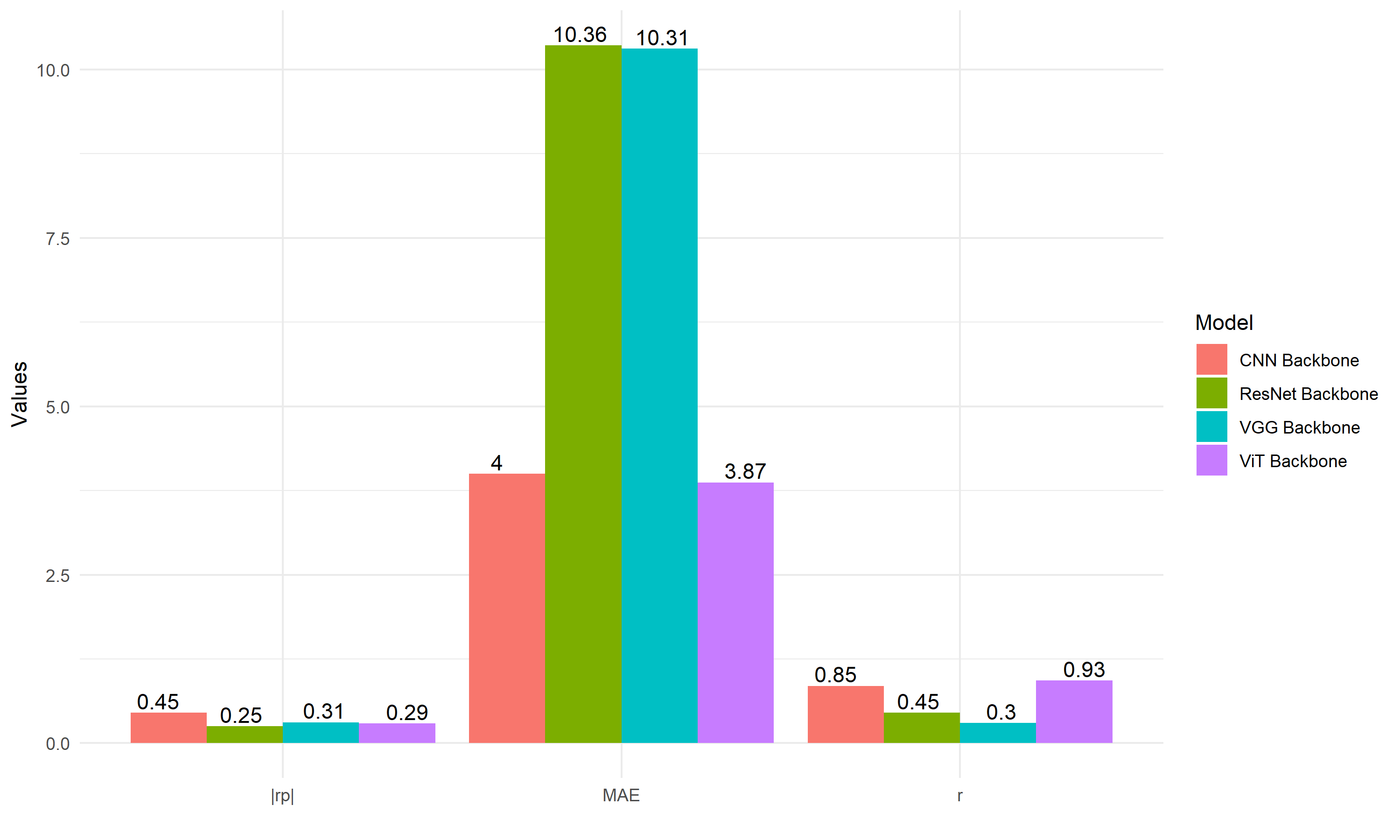}
\caption{The impact of the backbone architectures}
\label{figure 5}
\end{figure}

Continuing our ablation studies, we turned our focus to the backbone of Triamese-ViT. To assess the impact of different backbone architectures, we substituted the original ViT with alternative models like ResNet, a 5-layer CNN, and VGG19. These were then integrated with the Triamese MLP to evaluate how they influenced overall performance. The results of this experiment are detailed in Fig.~\ref{figure 5}.

For ease of interpretation, we chose to display the absolute value of the Spearman correlation coefficient between BAG and chronological age (denoted as $\left\|rp\right\|$) in the figure. A larger value of $\left\|rp\right\|$ indicates a stronger age bias in the model's predictions. According to our findings, the original ViT backbone proves to be the most effective for the Triamese structure. The 5-layer CNN also shows commendable adaptability, registering an MAE of 6, a Spearman correlation (r) of 0.85, and $\left\|rp\right\|$ of 0.45.

In stark contrast, ResNet and VGG19 appear significantly less suited for the Triamese framework. Both these architectures yielded MAEs exceeding 10, which are highly unfavorable outcomes for brain age estimation. This experiment underscores the importance of selecting an appropriate backbone model for the Triamese structure to ensure optimal performance.

Next, we explore the unique structures within our Triamese-ViT model, particularly focusing on the individual contributions of the three Vision Transformers (ViTs) oriented along different axes of the MRIs. These are the $ViT_{x}$ with dimensions (91,109,91), $ViT_{y}$ with dimensions (91,91,109), and $ViT_{z}$ with dimensions (109,91,91). The performance of each of these orientation-specific ViTs is crucial in understanding the efficacy of the combined Triamese-MLP structure. 

Additionally, we tested a variant model, $Triamese_{map}$, which also utilizes three ViTs on different viewpoints. However, unlike the standard Triamese-ViT, each ViT in $Triamese_{map}$ outputs a feature map from the Transformer Encoder, rather than a direct prediction from the MLP Head. The Triamese MLP in this variant then takes as input the concatenated feature maps from the three ViTs to make the final prediction.

The comparative performance of these models, including each individual orientation ViT and the $Triamese_{map}$ variant, is presented in Table ~\ref{Table 2}. This comparison is key to demonstrating the value added by the Triamese MLP in synthesizing the perspectives from the three distinct ViT orientations, highlighting the importance of integrating these viewpoints for more accurate brain age estimation.

\begin{table}[htbp]
    \centering
    \begin{tabular}{p{4cm}lll}
        \hline
        Algorithm & MAE & r & rp \\
        \hline
        Triamese-ViT & $\mathbf{3.87}$ & $\mathbf{0.93}$ & -$\mathbf{0.29}$ \\
        $ViT_{x}$   & 4.42& 0.78 & 0.33 \\
        $ViT_{y}$  & 4.99     & 0.92 & -0.29\\
        $ViT_{z}$ & 5.29     & 0.73 & -0.37\\
        $ViT_{map}$ & 5.04     & 0.61 &-0.55\\

        \hline
    \end{tabular}
    \caption{Unique structures performance}
    \label{Table 2}
\end{table}

Table ~\ref{Table 2} says Triamese MLP supports a great improvement of performance, for MAE, $ViT_{x}$ is the second best with 4.42, $ViT_{z}$ is the worst with 5.29. As for r, $ViT_{y}$ has the highest value with 0.92, This is closely followed by the combined Triamese-ViT model. Notably, $ViT_{map}$, which uses concatenated feature maps for prediction, shows the lowest correlation value at 0.61. Regarding the aspect of fairness, only $ViT_{map}$ displays a strong negative correlation. This suggests a significant reduction in age bias. Conversely, the other models, including the individual orientation-specific ViTs, exhibit minimal ageism in their predictions.

Overall, the data in Table 2 strongly supports the efficacy of the Triamese MLP in enhancing both the accuracy and fairness of brain age estimation, validating the design of our Triamese-ViT model.

\subsection{Explainable Results for Triamese-ViT}

\begin{figure*}[htbp]
\centering
\includegraphics[scale=0.8]{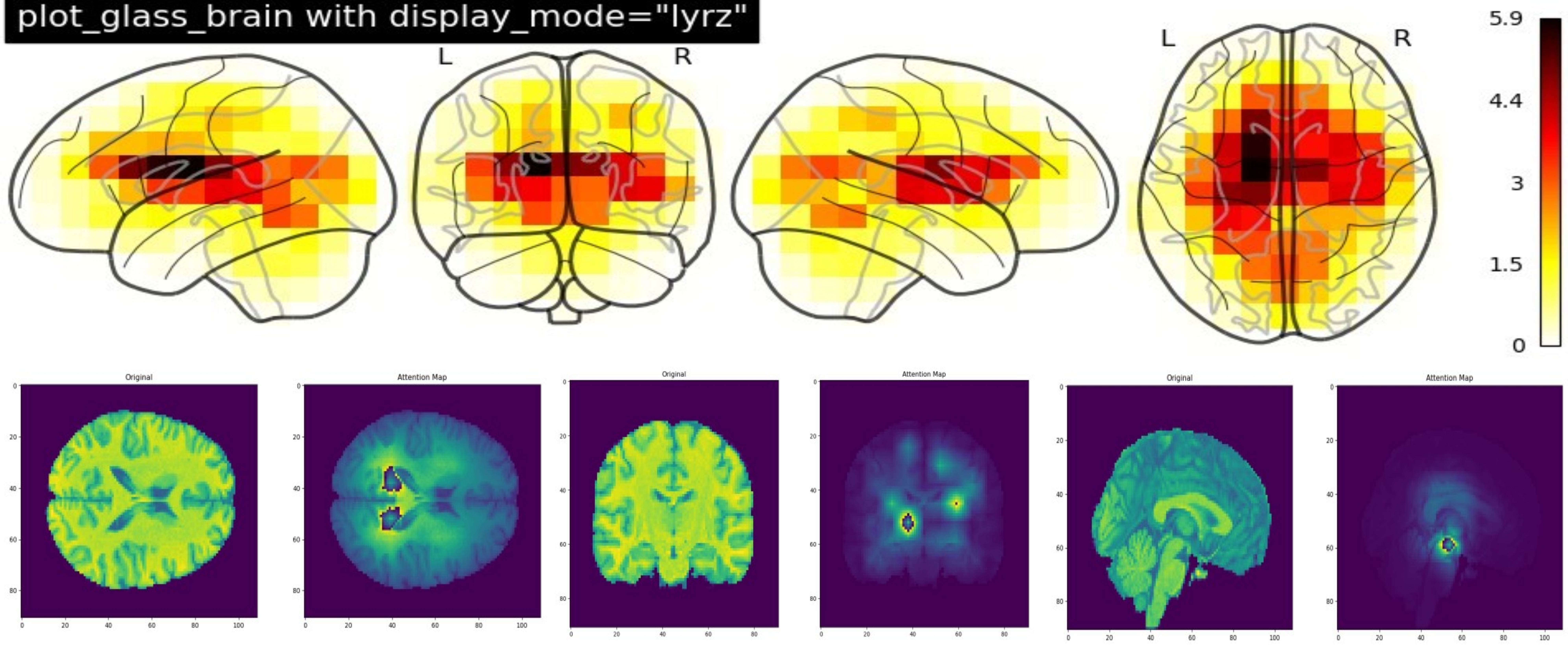}
\caption{Comparison between the attention map and occlusion analysis from Triamese-ViT. The upper half showcases the results from the occlusion analysis, while the lower half displays the results from the attention map. Both halves collectively highlight the specific regions of the brain that the Triamese-ViT model prioritizes and considers most informative for determining age.  }
\label{figure 7}
\end{figure*}

In this section, we delve into the explainable results generated by the Triamese-ViT model. Fig.~\ref{figure 7} displays outcomes obtained through two different methods: occlusion analysis and the attention map feature of the Triamese-ViT.

The upper half of the figure, representing the occlusion analysis, indicates that the removal of the Basal Ganglia and Thalamus significantly impacts the model’s predictions. This finding underscores the importance of these brain areas in age estimation. Additionally, the analysis suggests that the left side of the brain has a more pronounced effect on the results.

The lower half of the figure, showing the attention map results, offers more nuanced insights. It corroborates the significance of the Basal Ganglia and Thalamus but also highlights the Midbrain as a key influencer. This detailed analysis reaffirms the greater importance of the left brain in the prediction process.

Comparing these two methods, it is evident that the Triamese-ViT’s attention maps not only align with the findings from traditional explainable AI (XAI) methods but also provide additional, detailed insights.

Supporting this, various studies have linked critical brain functions and diseases to these regions. ~\cite{wang2022roles,xiong2022altered} have associated conditions like Parkinson's disease, Tourette's syndrome, Huntington's disease, Alzheimer's Disease, and addiction with the Basal Ganglia. ~\cite{russo2022age,smajic2022single} discuss the role of the substantia nigra in the midbrain, a dopamine-producing area crucial for movement regulation and significantly affected in Parkinson’s disease. Furthermore, the Thalamus, as noted in ~\cite{shine2023impact}, acts as a central hub for sensory information processing and plays a vital role in attention coordination.

These scientific findings validate the attention map’s emphasis on the Basal Ganglia, Midbrain, and Thalamus. Their crucial roles in various brain diseases directly relate to brain age estimation. Thus, the results from our Triamese-ViT not only demonstrate its strong capability for explainability but also have significant implications for understanding and studying brain diseases.

\section{Discussion}
In our research, we introduce a groundbreaking deep-learning algorithm, Triamese-ViT, used for brain age estimation. This model is benchmarked against other leading models in the field, demonstrating its remarkable superiority in performance. The pivotal contribution of Triamese-ViT is its innovative Triamese structure. This design is a first in the realm of brain age estimation, merging the benefits of global context understanding with detailed image analysis. It delves into the intricate relationships between image patches, leading to predictions that are not only more comprehensive and accurate but also highly interpretable. This blend of detailed analysis and contextual understanding sets Triamese-ViT apart, marking a significant advancement in the field of brain age estimation.

In this project, while traditional CNN-based models use complete 3D MRI scans for detailed predictions, their focus on small, localized pixels can miss crucial global features, affecting prediction accuracy, especially in complex brain structure analysis. Conversely, Vision Transformers (ViTs), adapted from natural language processing, enhance brain age estimation by dissecting images into patches and analyzing their interrelations, offering detailed insights. However, ViTs often overlook the overall image context and typically process 3D MRIs as 2D slices, which may result in the loss of important depth information.

Our innovation, Triamese-ViT, draws inspiration from the Siamese Network, which shows great performance in various fileds. Triamese-ViT is constructed with three ViTs, each analyzing 3D images from different viewpoints. This setup, combined with a Triamese MLP for feature extraction and prediction, effectively harnesses the strengths of both CNNs and ViTs while mitigating their respective weaknesses.

The ViT backbone enables detailed image analysis and understanding of inter-patch relationships. The Triamese structure, on the other hand, ensures a comprehensive assessment of the whole image from multiple perspectives, preserving the depth aspect in the estimation process.

Triamese-ViT, when tested on a public dataset, demonstrated excellent performance: a Mean Absolute Error (MAE) of 3.87, a 0.93 Spearman correlation with chronological age, and a -0.29 Spearman correlation between Brain Age Gap (BAG) and chronological age. These results signify not only high predictive accuracy but also a reduction in age bias, marking a notable advancement in brain age estimation.

Moreover, Triamese-ViT excels in interpretability, crucial in medicine. Its attention maps provide more detailed insights compared to occlusion analysis and are integrated into the prediction process, offering a faster, user-friendly interpretation. This feature is especially valuable in medical settings where swift, accurate decision-making is essential.

The attention maps generated by Triamese-ViT pinpoint the Basal Ganglia, Thalamus, and Midbrain as crucial areas for brain age estimation. These regions play a significant role in determining the health of a patient’s brain according to the model. Supporting this, several medical studies~\cite{shine2023impact,smajic2022single,russo2022age,xiong2022altered,wang2022roles} have established a strong correlation between these brain areas and various severe neurological diseases, underlining the model's robust interpretability.

Furthermore, both the attention maps and occlusion analysis consistently indicate a greater influence of the left brain in age estimation. This finding could be attributed to the larger proportion of right-handed individuals in the dataset, as right-handedness is often associated with more developed left-brain regions. This aspect of our findings opens avenues for further research and underscores the depth and reliability of the insights provided by Triamese-ViT, making it an invaluable tool for advancing our understanding of brain health and aging.

\section{Conclusion}
In this paper, we introduced Triamese-ViT, a novel deep-learning architecture applied to brain age estimation. Triamese-ViT exhibits exceptional accuracy, fairness, and interpretability, surpassing existing advanced algorithms in the field. Its high performance, low bias, and robust interpretability make it well-suited for medical research. The model's user-friendly nature enhances its applicability in clinical settings where efficiency and clarity are crucial. Triamese-ViT represents a meaningful contribution to the integration of AI in medicine, offering potential advancements in research and applications. We envision its utility not only in advancing brain age estimation but also as a valuable tool for broader medical AI research and development.

{
    \small
    \bibliographystyle{ieeenat_fullname}
    \bibliography{main}
}


\end{document}